\documentclass{article}

\usepackage{arxiv}

\usepackage[T1]{fontenc}    % use 8-bit T1 fonts
\usepackage{hyperref}       % hyperlinks
\usepackage{hyperref}
\hypersetup{
    colorlinks=true
}
\usepackage{url}            % simple URL typesetting
\usepackage{booktabs}       % professional-quality tables
\usepackage{amsfonts}       % blackboard math symbols
\usepackage{nicefrac}       % compact symbols for 1/2, etc.
\usepackage{microtype}      % microtypography
\usepackage{lipsum}

\usepackage{amsmath}

\usepackage[ruled]{algorithm2e}
\SetKwRepeat{Do}{do}{while}%
\usepackage{graphicx}
\usepackage{multirow}

\title{Gumbel-softmax Optimization: A Simple General Framework for Combinatorial Optimization Problems on Graphs}

\author{
  Jing Liu\\
  School of Systems Science\\
  Beijing Normal University\\
  \texttt{jing.liu@mail.bnu.edu.cn} \\
   \And
     Fei Gao\\
  School of Systems Science\\
  Beijing Normal University\\
  \texttt{Philip.sss@mail.bnu.edu.cn} \\
\And   
 Jiang Zhang\thanks{Use footnote for providing further
    information about author (webpage, alternative
    address)---\emph{not} for acknowledging funding agencies.}\\
  School of Systems Science\\
  Beijing Normal University\\
  \texttt{zhangjiang@bnu.edu.cn} \\
}

\begin{document}
\maketitle

\begin{abstract}
Many problems in real life can be converted to combinatorial optimization problems (COPs) on graphs, that is to find a best node state configuration or a network structure such that the designed objective function is optimized under some constraints. However, these problems are notorious for their hardness to solve because most of them are NP-hard or NP-complete. Although traditional general methods such as simulated annealing (SA), genetic algorithms (GA) and so forth have been devised to these hard problems, their accuracy and time consumption are not satisfying in practice. In this work, we proposed a simple, fast, and general algorithm framework called Gumbel-softmax Optimization (GSO) for COPs. By introducing Gumbel-softmax technique which is developed in machine learning community, we can optimize the objective function directly by gradient descent algorithm regardless of the discrete nature of variables. We test our algorithm on four different problems including Sherrington-Kirkpatrick (SK) model, maximum independent set (MIS) problem, modularity optimization, and structural optimization problem. High-quality solutions can be obtained with much less time consuming compared to traditional approaches.
\end{abstract}
% keywords can be removed
\keywords{Gumbel-softmax \and Combinatorial optimization problems \and Sherrington-Kirkpatrick model \and Maximum independent set}

\section{Introduction}
\label{section: intro}
Combinatorial optimization problems (COPs) ask for optimizing a given objective function whose domain is a discrete but large configuration space\cite{karp1972reducibility}. COPs arise in many disciplines and many real-world optimization problems can be converted to COPs. In computer science, there exist a large number of COPs defined on graphs, e.g., maximal independent set (MIS) and minimum vertex cover (MVC)\cite{karp1972reducibility}. In these problems, one is asked to give a largest (or smallest) subset of the graph under some constraints. In statistical physics, finding the ground state configuration of spin glasses model where the energy is minimized is another type of COPs\cite{mezard1987spin}. Many problems of network science can also be treated as COPs, for example, modularity optimization\cite{newman2006modularity} asks to specify which community one belongs to that maximize the modularity value. While, structural optimization asks to reveal the underlying network structure under some constraints. For example, finding the network structure to recover the observed network dynamics if the dynamical rules are given. This is also called network reconstruction problem\cite{timme2007revealing, casadiego2017model}. In general, the space of possible solutions of mentioned problems is typically very large and grows exponentially with system size, thus impossible to solve by exhaustion. 
 
Many methods have been proposed to solve COPs. Traditional general optimization methods such as simulated annealing (SA)\cite{kirkpatrick1983optimization}, genetic algorithm (GA)\cite{davis1991handbook}, extremal optimization (EO) \cite{boettcher2000nature} usually suffer from slow convergence and are limited to system size up to thousand. Although there exist many other heuristic solvers such as local search\cite{andrade2012fast}, they are usually domain-specific and require special knowledge.

Recently a great deal of effort has been devoted to applying machine learning to solve COPs. Khalil et al.\cite{khalil2017learning} used a combination of reinforcement learning and graph embedding called S2V-DQN to address combinatorial optimization problems on graphs. Li et al.\cite{li2018combinatorial} used graph convolution networks (GCNs)\cite{kipf2016semi} with classic heuristics and achieved state-of-art performance compared to S2V-DQN. However, these two algorithms all belong to supervised learning, thus contain two stages of problem solving: first training the solver and then testing. Although relatively good solutions can be obtained efficiently, it takes a long time for training the solver and the quality of solutions depends heavily on the quality and the amount of the data for training, which is hardly for large graphs. 
Some techniques have been developed in reinforcement learning area which seeks to solve the problems directly without training and testing stages. For example, REINFORCE algorithm is a typical gradient estimator for discrete optimization. However, it suffers from high variance and time-consuming problems. 
Recently reparameterization trick developed in machine learning community such as Gumbel-softmax\cite{gumbel, concrete} provides another approach for differentiable sampling. It allows us to pass gradients through samples directly and this method has not been applied to solving COPs on graphs so far.

In this work, we present a novel optimization method based on Gumbel-softmax, called Gumbel-softmax optimization (CSO) method. Under the naive mean field assumption, we  assume that the joint probability of a configuration is factorized by the product of marginal probabilities. The marginal probabilities, which stand for the states of nodes or edges, are parameterized by a set of learnable parameters. We can sample a configuration through Gumbel-softmax and compute the objective function. Then we run backpropagation algorithm and update parameters. Furthermore, GSO can be implemented in parallel easily on GPU to simultaneously solve the optimization problem under different initial conditions, and one solution with the best performance can be selected. We first introduce four different problem in Section \ref{sec:preliminary}. Then the details of GSO algorithm is explained in Section \ref{sec:method}. In Section \ref{sec:experiment} we introduce our main results on all four problems, and compare with different benchmark algorithms. The results show that GSO can achieved satisfying solutions and also benefit from time consumption. Finally some concluding remarks are given in Section \ref{sec:conclusion}.

\section{Preliminary}
\label{sec:preliminary}
The combinatorial optimization problems that we concerned can be formulated as follows. Let's consider $N$ variables $s_i$ where $i\in \{1,2,\cdots,N\}$ and each variable can take discrete values: $s_i\in \{1,2,\cdots,K\}$. The combination of the values for all variables is defined as a configuration or a state $\mathbf s=(s_1,s_2,\cdots,s_N)\in \{1,2,\cdots,K\}^N$. The objective function $E$ is a map between the configuration space and the real number field, $E\in \{1,2,\cdots,K\}^N\times \mathbb{R}$. Thus, the goal of the combinatorial optimization problem is to find a configuration $\mathbf s^*$ such that the objective function is minimized, i.e., $E(\mathbf s^*)\leq E(\mathbf s)$ for all $\mathbf s\in \{1,2,\cdots,K\}^N$. Usually, a set of constraints must be considered which can be formulated as boolean functions on $\mathbf s$. Nevertheless, they can be absorbed into the objective functions by introducing the punishment parameter. Therefore, we only consider the optimization problem without constraints in this paper for the simplicity.

Here, we consider two kinds of COPs on graphs: one is node optimization problem, the other is structure optimization problem. The node optimization problem can be stated as follows. Given a graph $\mathcal G = (\mathcal V, \mathcal E)$ where $\mathcal V = \{v_i\}_{i=1:N}$ is the set of $N=|\mathcal V|$ vertices in $\mathcal G$ and $\mathcal E$ is the set of $M = |\mathcal E|$ edges. Each vertex is a variable $s_i \in \{1, 2, \cdots, K\}$ in the optimization. 

The structure optimization problem is to find an optimal network structure such the objective function defined on graphs can be optimized. If we treat the adjacency matrix of the graph as the state of the system, and each entry of the matrix is a variable which taking value from $\{0,1\}$, then structural optimization problem is also a COP.

In this work we focus on four typical optimization problems:

\paragraph{Sherrington-Kirkpatrick (SK) model} SK model\cite{sherrington1975solvable} is a celebrated spin glasses model defined on a complete graph where all vertices are connected. The objective function, or the ground state energy of this model is defined as
\begin{equation}
    E(s_1, s_2, \cdots, s_N) = -\sum_{1\leq i < j \leq N} J_{ij} s_i s_j,
    \label{eq:skmodel}
\end{equation}
where $s_i \in \{-1, +1\}$ and $J_{ij} \sim \mathcal{N}(0, 1/N)$ is the coupling strength between two vertices. We are asked to give a configuration $\mathbf s = (s_1, s_2, \cdots, s_N)$ where the ground state energy is minimized. The number of possible configurations is $2^N$ and in fact, finding the ground state configuration of SK model is an NP-hard problem\cite{mezard1987spin}.

\paragraph{Maximal Independent Set (MIS)} MIS problem is another well-known NP-hard problem. The reason why we focus on MIS problem is that many other NP-hard problems such as Minimum Vertex Cover (MVC) problem and Maximal Clique (MC) problem can be reduced to an MIS problem. An MIS problem asks to find the largest subset $\mathcal V^{\prime} \subseteq \mathcal V$ such that no two vertices in $\mathcal V^{\prime}$ are connected by an edge in $\mathcal E$. The Ising-like objective function consists of two parts: 
\begin{equation}
    E(s_1, s_2, \cdots, s_N) = -\sum_i s_i + \alpha \sum_{ij\in \mathcal E}s_i s_j,
    \label{eq:mis}
\end{equation}
where $s_i \in \{0, 1\}$ and the second term provides an penalty for two connected vertices being chosen. $\alpha$ is the hyperparameter for the strenghth of the punishment. The first term is the number of vertices in $\mathcal V^{\prime}$. 

\paragraph{Modularity optimization} Modularity is a graph clustering index for detecting community structure in complex networks\cite{fortunato2010community}. Many modularity optimization methods, e.g., hierarchical agglomeration\cite{newman2004fast}, spectral algorithm\cite{newman2006modularity} and extremal optimization\cite{duch2005community} have been proposed and it is suggested that maximizing modularity is strongly NP-complete\cite{brandes2006maximizing}. In general cases where a graph is partitioned into $K$ communities, the objective is to maximize the following modularity $Q$:
\begin{equation}
E(s_1, s_2, \cdots, s_N) = \frac{1}{2M}\sum_{ij}\left[A_{ij}-\frac{k_{i} k_{j}}{2 M}\right]\delta(s_i, s_j),
\label{eq:modularity}
\end{equation}
% Q=\frac{1}{2 M} \sum_{i j} \sum_{r}\left[A_{i j}-\frac{k_{i} k_{j}}{2 M}\right] S_{i r} S_{j r}=\frac{1}{2 M} \Tr\left(\mathbf{S}^{\mathrm{T}} \mathbf{B} \mathbf{S}\right)
where $k_i$ is the degree of vertex $i$, $s_i\in \{1,2,\cdots,K\}$, where $K$ is the number of communities, $\delta(s_i,s_j)=1$ if $s_i=s_j$ and $0$ otherwise.

\paragraph{Structural optimization}
The problem of structural optimization is to find an optimal network structure for some objective function defined on a network\cite{eusuff2003optimization, boyce1973optimal}. This problem can also be considered as a special example of a COP if the element of the adjacency matrix $\mathbf a_{ij}\in \{0,1\}$ is the optimized variable, and a configuration $\mathbf A\in \{0,1\}^{N\times N}$ is the adjacency matrix of a potential graph, where $N$ is the number of nodes in the network which is a fixed value.

For example, we can find an optimal network to fit the observed node dynamics, in which, we are asked to optimize the network structure $\mathbf A$ to match the given observed time series data of all nodes $\{\mathbf x_1, \mathbf x_2, \cdots, \mathbf x_T\}$, where $\mathbf x_t$ is the node state vector at $t$, and $T$ is the length of the time window for our observation. The network dynamical rule $\mathcal{F}$ such that $\mathbf x_{t+1}=\mathcal{F}(\mathbf x_t, \tilde{A})$ is known but the underlying real network structure $\tilde{A}$ is unknown for us. Our task is to recover the real network structure $\mathbf{\tilde{A}}$ by minimizing the following objective function:

\begin{equation}
    E(\mathbf A)=\sum_{t=2}^{T}(\mathbf x_t-\mathcal{F}(\mathbf x_{t-1},\mathbf A))^2.
    \label{eq:net-recons}
\end{equation}

Conventionally, this problem is also called network reconstruction problem in \cite{timme2007revealing,casadiego2017model}. 

\section{Methodology}
\label{sec:method}
We will show that one particular computational approach can solve all the problems mentioned above in a unified way. Notice that although the objective functions in Equation \ref{eq:skmodel}, \ref{eq:mis}, and \ref{eq:modularity} all have Ising-like form, Equation \ref{eq:net-recons} is much more complicated than others because the dynamical rule $\mathcal{F}$ may be very complex and even non-analytic. However, we will show that our computational framework can work on all of them.

\paragraph{Naive mean field approximation} 
The basic idea of our method is to factorise the possible solutions into $N$ independent Bernouli variables, this is called naive mean field approximation in statistical physics. We assume that vertices in the network are independent and the joint probability of a configuration $\mathbf s = (s_1, s_2, \cdots, s_N)$ can be written as a product distribution\cite{wainwright2008graphical}:
\begin{equation}
    \Pr(s_1, s_2, \cdots, s_N) = \prod_{i=1}^N \Pr(s_i).
    \label{eq:jointdistribution}
\end{equation}
The marginal probability $\Pr(s_i)\in [0,1]^K$ can be parameterized by a set of parameters $\theta_i$ which is easily generated by Sigmoid or softmax function. 

We sample the possible solution $\mathbf s$ according to Equation \ref{eq:jointdistribution}, and evaluate the objective function $E(\mathbf s;\pmb \theta)$ under the given parameters. Then by using the automatic differential techniques empowered by current popular deep learning frameworks such as PyTorch\cite{paszke2017automatic} and TensorFlow\cite{abadi2016tensorflow}, we can obtain the gradient vector $\partial E(\mathbf s;\pmb \theta) / \partial \pmb \theta$ for all parameters of variables. And the parameters $\pmb \theta$ will be updated according to the gradient vector.
However, the sampling process in the above procedure is not differentiable which is required in all deep learning frameworks to compute gradient vector because the sampling operation introduces stochasticity. 

\paragraph{Gumbel-softmax technique}  Traditionally we can resort to Monte Carlo gradient estimation techniques such as REINFORCE\cite{williams1992simple}. Gumbel-softmax\cite{gumbel}, a.k.a. concrete distribution\cite{concrete} provides an alternative approach to tackle the difficulty of non-differentiability. Consider a categorical variable $s_i$ that can take discrete values $s_i \in \{1,2,\cdots, K\}$. This variable $s_i$ can be parameterised as a \textit{K}-dimensional vector $(p_1, p_2, \cdots, p_K)$ where $\theta_i$ is the probability that $\theta_i=\Pr(s_i=r), r=1, 2, \cdots, K$. Instead of sampling a hard one-hot vector, Gumbel-softmax technique give a \textit{K}-dimensional sample vector where the \textit{i}-th entry is
\begin{equation}
    \hat{p}_{i}=\frac{\exp \left(\left(\log \left(p_{i}\right)+g_{i}\right) / \tau\right)}{\sum_{j=1}^{K} \exp \left(\left(\log \left(p_{j}\right)+g_{j}\right) / \tau\right)} \quad \text { for } i=1,2, \cdots, K,
\label{eq:gumbel}
\end{equation}
where $g_i \sim \text{Gumbel}(0,1)$ is a random variable following standard Gumbel distribution and $\tau$ is the temperature parameter. Notice that as $\tau \rightarrow 0$, the softmax function will approximate $\text{argmax}$ function and the sample vector will approach a one-hot vector. And the one-hot vector can be regarded as a sample according to the distribution $(p_1,p_2,\cdots,p_K)$ because the unitary element will appear on the $i^{th}$ element in the one-hot vector with probability $p_i$, therefore, the computation of gumbel-softmax function can simulate the sampling process. Furthermore, this technique allows us to pass gradients directly through the ``sampling'' process because all the operations in Equation \ref{eq:gumbel} are differentiable. In practice, it is common to adopt a annealing schedule from a high temperature $\tau$ to a small temperature. 

\begin{figure}[!htbp]
	\centering
	\includegraphics[width=0.75\linewidth]{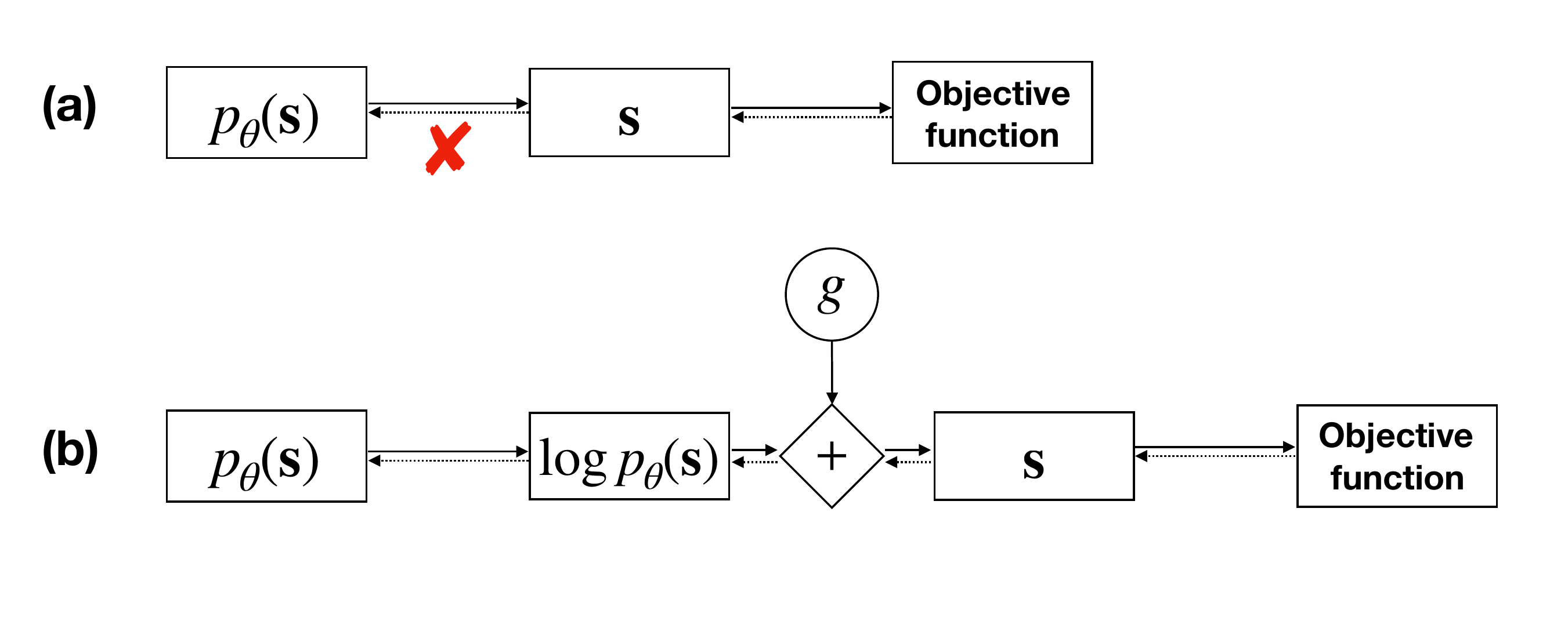}
	\caption{In (a), we cannot use backpropagation algorithm because the sampling from $p_{\theta}(\mathbf s)$ is stochastic and we cannot compute the gradients. In (b), however, due to the Gumbel-softmax technique, the gradients can flow from $\mathbf s$ to $p_{\theta}(\mathbf s)$. }
	\label{fig:arch}
\end{figure}

\paragraph{Gumbel-softmax optimization}
By introducing the Gumbel-softmax technique, the whole process of optimization solution can be stated as follows (also see Figure \ref{fig:arch}):
\begin{enumerate}
\item Initialization for $N$ vertices: $\pmb \theta = (\theta_1, \theta_2, \cdots, \theta_N)$.
\item Sample from $\Pr (s_i)$ simultaneously via gumbel-softmax technique and then calculate the objective function.
\item Backpropagation to compute gradients $\partial E(\mathbf s; \pmb \theta)/\partial \pmb \theta$ and update parameters $\pmb \theta = (\theta_1, \theta_2, \cdots, \theta_N)$ by gradient descent. 
\end{enumerate}

We point out that our method can be implemented in parallel on GPU: we can simultaneously initialize $N_{\text{bs}}$ different initial values and calculate $N_{\text{bs}}$ objective functions. When the training procedure is finished, we select the result with best performance from $N_{\text{bs}}$ candidates. We present our proposed method in Algorithm \ref{algorithm}.

\begin{algorithm}[H]
% \SetAlgoLined
\KwIn{Problem size $N$, batch size $N_{\text{bs}}$, learning rate $\eta$, and Graph $\mathcal{G=}$ for node state optimization problem or dynamical rule $\mathcal{F}$ and node observations $(\mathbf{x_1},\mathbf{x_2},\cdots,\mathbf{x_T})$ for structural optimization problem.}
\KwOut{solution with the best performance}
Initialize $\pmb \theta = (\theta_1, \theta_2, \cdots, \theta_N) \in \mathbb R^{N_{bs}\times N \times K}$\;
\Repeat{Convergence}
{
$\mathbf{s} \leftarrow \text{Gumbel-softmax sampling from } p_{\pmb \theta}$\;
$E \leftarrow E(\mathbf s;\pmb \theta$)\;
Backpropagation\;
$\pmb{\theta} \leftarrow \pmb{\theta} - \eta \frac{\partial E}{\partial \pmb{\theta}}$\;
}
Select solution with the best performance\;
\caption{Gumbel-softmax Optimization (GSO)}
\label{algorithm}
\end{algorithm}

Notice that to parallelize the algorithm on GPU, $\mathbf{s}\in \{1,2,\cdots,K\}^{N_{bs}\times N}$ and $E\in \mathbb{R}^{N_{bs}}$, and their first dimensions are all batch dimensions.
\section{Experiment}
\label{sec:experiment}
\paragraph{Experimental settings} We conduct experiments on all four problems described in Section \ref{sec:preliminary}. For SK model, we optimize the ground
state energy with various sizes ranging from 256 to 8192. For MIS
problem, we use citation network datasets: \emph{Cora, Citeseer and PubMed} and treat them as undirected
networks. For modularity optimization, we use four real-world datasets
studied in \cite{newman2004fast, newman2006modularity, duch2005community}: \emph{Zachary, Jazz, C.elegans and E-mail.} We
compare our proposed method to other classical optimization methods and state-of-the-art deep learning approaches, e.g., (1) simulated annealing (SA)\cite{kirkpatrick1983optimization}: a general
optimization method inspired by Metropolis-Hastings algorithm; (2)
Extremal optimization (EO)\cite{boettcher2000nature}: a heuristic designed to address combinatorial
optimization problems; (3) Structure2Vec Deep Q-learning (S2V-DQN)\cite{khalil2017learning}: a reinforcement learning method to
address optimization problems over graphs; (4) GCNs\cite{li2018combinatorial}: a supervised
learning method based on graph convolutional networks (GCNs).

In practice, we choose the batch size $N_{\text{bs}}=128$. During optimization, we use Gumbel-softmax estimator to generate continuous
samples to calculate the objective function but discrete samples for the final output. We use Adam optimizer\cite{kingma2014adam} with different learning rate $\{\mathrm{1e-3, 1e-2, 1e-1, 1}\}$, annealing rate $\{0.99, 0.999, 0.9999\}$ and report results with best performance.

As for the task of structural optimization, we choose Coupled Map Lattice on a given but unknown graph $\tilde{A}$ as our dynamical rule, which is widely used to study the chaotic dynamics of spatially extended systems: 
\begin{equation}
x_{t+1}(i)=(1-s) f\left(x_{t}(i)\right)+\frac{s}{\operatorname{deg}(i)} \sum_{j \in \text{neighbor}(i)} f\left(x_{t}(j)\right)
\end{equation}
where $s$ is the coupling constant, $\text{deg}(i)$ is the degree of node $i$ and $neighbor(i)$ is the set of neighbor nodes of $i$ on graph $\tilde{A}$. Here, $f()$ is the logistic map function $f(x)=\lambda x(1-x)$. We conduct our experiments on random 4-regular graphs. We first sample an adjacency matrix $\mathbf A^{\prime}$ and then estimate the error of dynamics according to Eq \ref{eq:net-recons}.

\paragraph{Performance}
\begin{table}[!htbp]
\centering
\caption{\textbf{Results on optimization of ground state energy of SK model for different system size $N$ and instances $I$.} \cite{boettcher2005extremal} only reported results of system size up to $N=1024$. In the implementation of simulated annealing, the program failed to finish within 96 hours for $N=8192$. We also present the comparison of time consumption.}
\label{tab:sk-1}
\begin{tabular}{@{}cccccccc@{}}
\toprule[2pt]
\multirow{2}{*}{N} & \multirow{2}{*}{I} & \multicolumn{2}{c}{EO\cite{boettcher2005extremal}}             & \multicolumn{2}{c}{SA} & \multicolumn{2}{c}{GSO ($N_{bs}=1$)} \\ \cmidrule(l){3-8} 
                   &                    & $E_0$                & time        & $E_0$       & time (s) & $E_0$                  & time (s)         \\ \midrule[1pt]
256                & 5000               & \textbf{-0.74585(2)} & $\sim 268$s & -0.7278(2)  & 1.28     & -0.7270(2)             & \textbf{0.75}    \\
512                & 2500               & \textbf{-0.75235(3)} & $\sim 1.2$h & -0.7327(2)  & 3.20     & -0.7403(2)             & \textbf{1.62}    \\
1024               & 1250               & \textbf{–0.7563(2)}  & $\sim 20$h  & -0.7352(2)  & 15.27    & -0.7480(2)             & \textbf{3.54}    \\
2048               & 400                & -                    & -           & -0.7367(2)  & 63.27    & \textbf{-0.7524(1)}    & \textbf{5.63}    \\
4096               & 200                & -                    & -           & -0.73713(6) & 1591.93  & \textbf{-0.7548(2)}    & \textbf{8.38}    \\
8192               & 100                & -                    & -           & -           & -        & \textbf{-0.7566(4)}    & \textbf{26.54}   \\ \bottomrule[2pt]
\end{tabular}
\end{table}

\begin{table}[!htbp]
\centering
\caption{\textbf{Results on optimization of ground state energy of SK model for different system size $N$ and instances $I$.} We use different optimizer in gradient descent (GD). We also present the results of parallel version of our proposed method  where we choose $N_{bs}=128$.}
\label{tab:sk-2}
\begin{tabular}{@{}cccccccccc@{}}
\toprule[2pt]
\multirow{2}{*}{N} & \multirow{2}{*}{I} & \multicolumn{2}{c}{GD (Adam)} & \multicolumn{2}{c}{GD (L-BFGS)} & \multicolumn{2}{c}{GSO ($N_{bs}=1$)} & \multicolumn{2}{c}{GSO ($N_{bs}=128$)} \\ \cmidrule(l){3-10} 
                   &                    & $E_0$          & time (s)     & $E_0$        & time (s)         & $E_0$                & time (s)           & $E_0$                    & time (s)         \\ \midrule[1pt]
256                & 5000               & -0.6433(3)     & 2.84         & -0.535(2)    & 2.29             & -0.7270(2)           & 0.75               & \textbf{-0.7369(1)}      & \textbf{0.69}    \\
512                & 2500               & -0.6456(3)     & 2.87         & -0.520(3)    & 2.56             & -0.7403(2)           & 1.62               & \textbf{-0.7461(2)}      & \textbf{1.61}    \\
1024               & 1250               & -0.6466(4)     & 3.22         & -0.501(5)    & \textbf{2.73}    & -0.7480(2)           & 3.54               & \textbf{-0.7522(1)}      & 4.09             \\
2048               & 400                & -0.6493(2)     & 3.53         & -0.495(8)    & \textbf{3.06}    & -0.7524(1)           & 5.63               & \textbf{-0.75563(5)}     & 12.19            \\
4096               & 200                & -0.6496(5)     & 4.62         & -0.49(1)     & \textbf{3.55}    & -0.7548(2)           & 8.38               & \textbf{-0.75692(2)}     & 39.64            \\
8192               & 100                & -0.6508(4)     & 16.26        & -0.46(2)     & \textbf{4.82}    & -0.7566(4)           & 26.54              & \textbf{-0.75769(2)}     & 204.26           \\ \bottomrule[2pt]
\end{tabular}
\end{table}

Table \ref{tab:sk-1} and \ref{tab:sk-2} show the results of different approaches on the task of optimizing ground state of energy of SK model of sizes 256, 512, 1024, 2048, 4096 and 8192. Note that: 
(1) Although extremal optimization (EO) has obtained better results, this method is limited to system size up to 1024 and is extremely time-consuming.
(2) Under product distribution, the gradients of ground state energy can be obtained explicitly: from $\Pr(s_i = +1) + \Pr(s_i = -1) = 1$ and $\Pr(s_i = +1) \times (+1) + \Pr(s_i = -1) \times (-1) = m_i$ where $m_i$ is the average magnetization, we have:
\begin{equation}
\begin{aligned}
 \langle E \rangle&= -\sum_{1\leq i < j \leq N} J_{ij} m_i m_j\\ &= -\sum_{1\leq i < j \leq N} J_{ij} [2\Pr(s_i=+1)-1] [2\Pr(s_j=+1)-1]
\end{aligned}
\end{equation}
We can optimize this objective function through gradient descent (GD) with different optimizer, e.g. Adam or L-BFGS but the results are not satisfying. The reason is that there are too many frustrations and local minima in the energy landscape.
(3) With the implementation of the parallel version, the results can be improved greatly.
% ########################################################################

% ########################################################################
\begin{table}[!htbp]
\centering
\caption{\textbf{Results on MIS problems.}}
\label{tab:mis}
\begin{tabular}{@{}ccccccc@{}}
\toprule[2pt]
Graph    & size  & S2V-DQN\cite{khalil2017learning} & GCNs\cite{li2018combinatorial}          & GD (L-BFGS) & Greedy         & GSO      \\ \midrule[1pt]
Cora     & 2708  & 1381    & \textbf{1451}  & 1446        & \textbf{1451}  & \textbf{1451} \\
Citeseer & 3327  & 1705    & \textbf{1867}  & 1529        & 1818           & 1802          \\
PubMed   & 19717 & 15709   & \textbf{15912} & 15902       & \textbf{15912} & 15861         \\ \bottomrule[2pt]
\end{tabular}
\end{table}

Table \ref{tab:mis} presents the performance of MIS problem on three citation networks.
Our proposed method has successfully found the global optimal solution
on Cora dataset and obtained much better results compared to the sophisticated S2V-DQN\cite{khalil2017learning} on all three dataset. Although our results are not competitive with \cite{li2018combinatorial}, we must stressed that it is a supervised learning
algorithm. Besides, it also adopts graph reduction techniques and a parallelized local search algorithm. Our method, however, requires none of these tricks.
% ########################################################################

% ########################################################################
\begin{table}[!htbp]
	\centering
	\caption{\textbf{Results on modularity optimization.} We report the maximum modularity  and the corresponding number of communities.}
	\label{tab:mod}
	\begin{tabular}{@{}cccccccc@{}}
		\toprule[2pt]
		\multirow{2}{*}{Graph} & \multirow{2}{*}{size} & \multicolumn{2}{c}{\cite{newman2006modularity}} & \multicolumn{2}{c}{EO\cite{boettcher2005extremal}} & \multicolumn{2}{c}{GSO} \\ \cmidrule(l){3-8} 
		&  & $Q$ & No. comms & $Q$ & No. comms & $ Q $ & No. comms \\ \midrule[1pt]
		Zachary & 34 & 0.3810 & 2 & 0.4188 & 4 & \textbf{0.4198} & 4 \\
		Jazz & 198 & 0.4379 & 4 & \textbf{0.4452} & 5 & 0.4451 & 4 \\
		C. elegans & 453 & 0.4001 & 10 &\textbf{ 0.4342}  & 12 & 0.4304 & 8 \\
		E-mail & 1133 & 0.4796  & 13 & \textbf{0.5738}  & 15 & 0.5275 & 8 \\ \bottomrule[2pt]
	\end{tabular}
\end{table}

\begin{figure}[!htbp]
    \centering
    \includegraphics[width=0.5\linewidth,trim=50 50 50 80, clip]{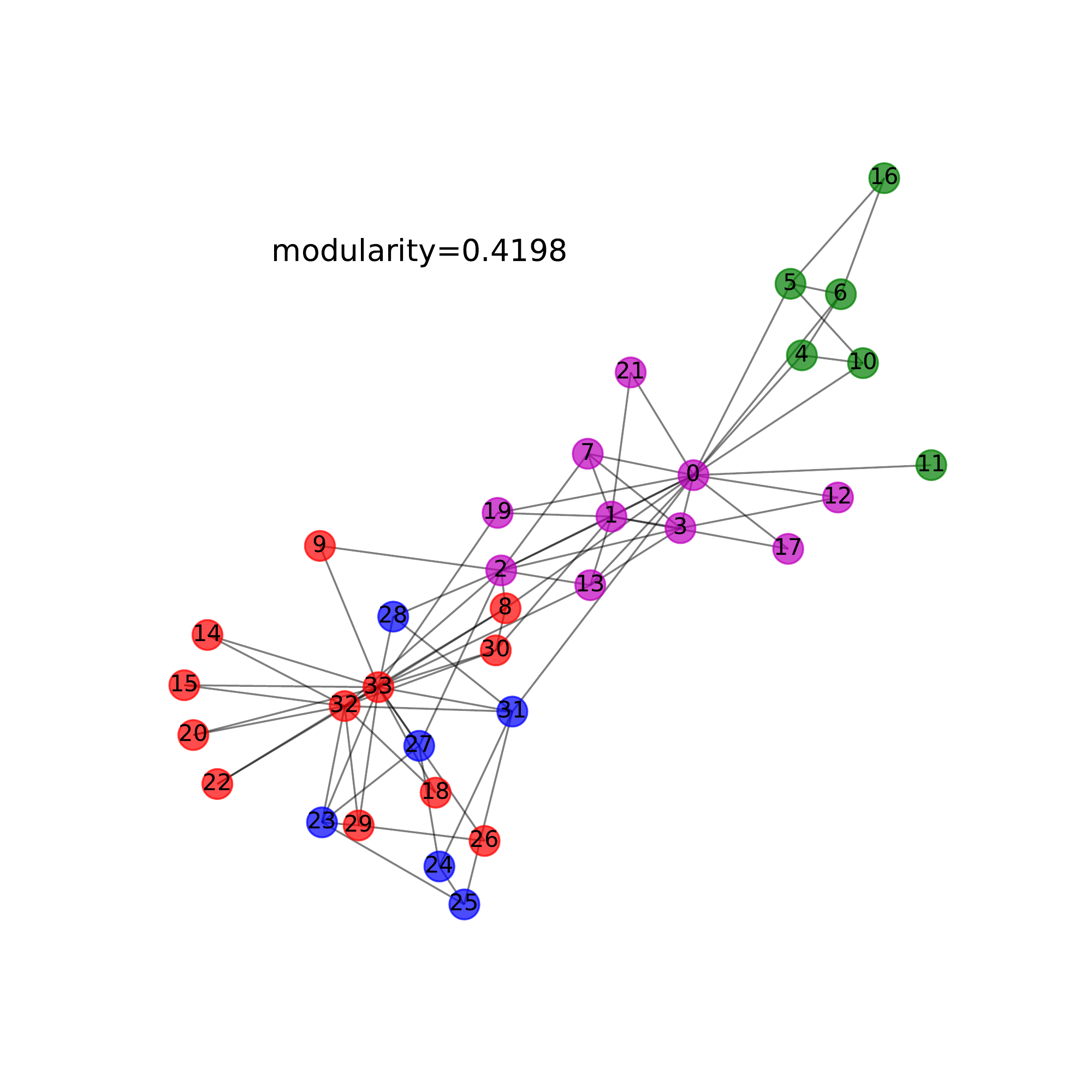}
    \caption{\textbf{Partition result on Zachary network.} The network is divided into four communities with modularity $Q=0.4198$.}
    \label{fig:zachary}
\end{figure}

On the task of modularity optimization, our method also obtained satisfying results, as shown in Table \ref{tab:mod}. Comparing to hierarchical agglomeration\cite{newman2004fast}, our proposed method has achieved much higher modularity on all datasets. In Figure \ref{fig:zachary} we present the partition obtained by our GSO method. 
% ########################################################################

% ########################################################################
\begin{table}[!htbp]
\centering
\caption{\textbf{Results on structural optimization.}}
\label{tab:net-recon}
\begin{tabular}{@{}ccc@{}}
\toprule[2pt]
N  & $s$, $\lambda$ & Accuracy (\%) \\ \midrule[1pt]
10 & 0.2, 3.5       &           96.80    \\
10 & 0.2, 3.8       &       100        \\
30 & 0.2, 3.5       &          93.11     \\
30 & 0.2, 3.8       &        100       \\ \bottomrule[2pt]
\end{tabular}
\end{table}

Here we present our results on the task of structural optimization in Table \ref{tab:net-recon}. We set coupling constant $s=0.2$ fixed. Because $r \approx 3.56995$ is the onset of chaos in the logistic map function, we choose $r=3.5$ and $r=3.8$ to present non-chaotic and chaotic dynamics respectively. For a random 4-regular graph with 10 and 30 nodes, our GSO method has obtained very good reconstruction accuracy and the performance obtained on chaotic dynamics is better than that on non-chaotic dynamics.

% ########################################################################

\paragraph{Time consumption}
% ########################################################################
\begin{figure}[!htbp]
	\centering
	\includegraphics[width=\linewidth]{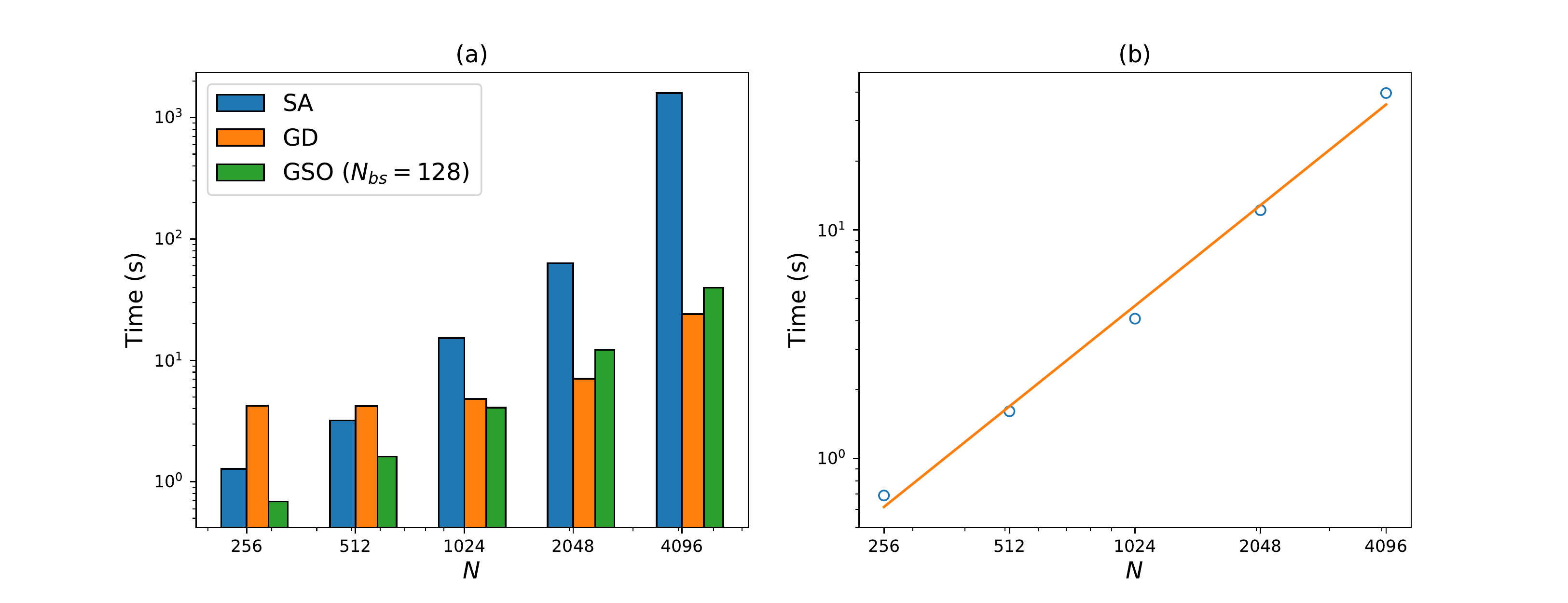}
	\caption{\textbf{Time consumption of GSO compared to other approaches. } (a): The time for simulated annealing (SA), gradient descent (GD) and our proposed GSO method on optimization of ground state energy of SK model. (b) A log-log plot of time versus system size $N$ and the slope is $1.46$. }
	\label{fig:time}
\end{figure}

Table \ref{tab:sk-1}, \ref{tab:sk-2} and Figure \ref{fig:time} shows the time consumption of different methods on SK model. We compare our proposed method with two baselines: gradient descent (GD) with Adam optimizer and simulated annealing (SA). Although gradient descent takes the least amount of time, its performance is the worst. Compared to simulated annealing, our method takes advantages in both performance and time consumption. As for extremal optimization, it is reported in \cite{boettcher2005extremal} that it takes nearly 20 hours for $N=1024$ and the algorithmic cost is $\mathcal O (N^4)$, which is extremely inefficient and time-consuming. We also plot the dependence of time with system size $N$ in Figure \ref{fig:time}. The slope of the log-log plot of time versus $N$ is 1.46, which indicates that the algorithmic cost is less than $\mathcal{O} (N^2)$. 

% ########################################################################

\section{Conclusion}
\label{sec:conclusion}
In this work, we have presented a novel optimization method, Gumbel-softmax optimization (CSO), for solving combinatorial optimization problems on graph. Our method is based on Gumbel-softmax, which allows the gradients passing through samples directly. We treat all vertices in the network are independent and thus the joint distribution is factorized as a product distribution, which enables Gumbel-softmax sample efficiently.
Our experiment results show that our method has good performance on all four tasks and also take advantages in time complexity. Comparing to traditional general optimization method, GSO can tackle large graphs easily and efficiently. Though not competitive to state-of-the-art deep learning based method, GSO have the advantage of not requiring neither training the solver nor specific domain knowledge. In general, it is an efficient, general and lightweight optimization framework for solving combinatorial optimization problems on graphs.

However, there is much space to improve our algorithm on accuracy. In this paper, we take the naive mean field approximation as our basic assumption, however, the variables are not independent on most problems. Therefore, much more sophisticated variational inference must considered in the future. Another way to improve accuracy is to combine other skills such as local search in our framework. 

\bibliographystyle{unsrt}  

\end{document}